%% file: root.tex
\documentclass[letterpaper, 10 pt, conference]{ieeeconf}  

\IEEEoverridecommandlockouts                              

\usepackage{graphics} 
\usepackage{epsfig} 
\usepackage{mathptmx} 
\usepackage{times} 
\usepackage{amsmath} 
\usepackage{amssymb}  
\usepackage{hyperref}
\usepackage{booktabs}

\usepackage{xcolor}

\title{\LARGE \bf
DAWN: Diffusion-based Alignment and unfastening Workflow Network
}

\author{
Hemant Kumar,
Piyush Goenka,
and Nikhil Chopra \\
\small Department of Mechanical Engineering\\
\small University of Maryland, College Park, MD, USA
\thanks{This work was supported by DEVCOM Army Research Laboratory through agreement number W911NF-22-2-0040.}%
}

\begin{document}

\maketitle
\thispagestyle{empty}
\pagestyle{empty}

\input{sections/abstract}

\input{sections/introduction}
\input{sections/litreview}

\input{sections/background}
\input{sections/methodology}
\input{sections/experiments}
\input{sections/conclusion}

\bibliographystyle{ieeetr}
\bibliography{references/references}

\end{document}

%% file: sections/abstract.tex
\begin{abstract}

We present DAWN (Diffusion-based Alignment and unfastening Workflow Network), a hierarchical robotic framework designed to perform real-world, unfastening tasks through multi-modal sensory integration. Autonomous robotic unfastening is crucial for improving recycling processes, enhancing safety, and reducing operational costs; yet, it remains predominantly manual due to inherent challenges in precise screwdriver engagement with fasteners, detailed force execution, and robustness to environmental variability. To address the complexity of learning long-horizon, contact-rich manipulation tasks, we subdivide the unfastening task procedure into two sequential stages: (1) alignment \& insertion, and (2) unfastening. DAWN consists of three key components: AlignNet, a diffusion-based policy ensuring precise screwdriver tip engagement; LatchNet, a multi-modal classifier reliably detecting successful stage transitions; and UnfastNet, which achieves stable, force-controlled screw removal. Experimental validation on a Franka Emika robot demonstrates that DAWN achieves alignment \& insertion success rates comparable to trained human operators, robustly transitions between stages, and attains an impressive success rate in unfastening across various unseen scenarios. Ablation studies further elucidate critical input modalities and effective training strategies, highlighting DAWN's scalability, robustness, and data efficiency.

\end{abstract}

%% file: sections/introduction.tex
\section{INTRODUCTION}

In recent years, the rapid growth in electronics and automotive waste poses significant environment and economic challenges, driving demand for efficient recycling and remanufacturing solutions \cite{wang2023energy}. Traditional disassembly methods remain predominantly manual and labor-intensive, exposing human workers to significant risks from hazardous materials, such as high-voltage battery components \cite{davis2019strong}. Automating the disassembly process can significantly reduce labor costs, improve efficiency, enhance safety by minimizing human involvement, and recovering high-value components for reuse \cite{yildiz2019dcnn, zhang2023novel, zhou2022design, zhou2023you}. Robotic automation presents a transformative solution capable of addressing these critical challenges, fundamentally reshaping the sustainability and profitability of recycling and remanufacturing industries.

Unfastening is a fundamental task in the disassembly process, accounting for anywhere between 39.9\% to 60\% of the total disassembly time, depending on the product category \cite{liu2020collaborative, peeters2016tool}. Due to its high labor requirement, unfastening has become a focal point for research on automation methods. Successful robotic unfastening demands precise engagement of the screwdriver tip with the screw head, detailed force and motion control, and robustness to environmental variations. Despite advancements in vision-based screw detection and localization \cite{yildiz2019dcnn, foo2021screw, zhang2023automatic, al2022machine, klas2021kit, deng2022tiny}, real-world disassembly settings present persistent difficulties: significant inaccuracies during visual identification and localization.
Moreover, transforming vision-based detections into accurate three-dimensional coordinates is error-prone, often necessitating manual intervention \cite{zhou2024towards}. Furthermore, collecting extensive annotated datasets to train reliable vision-based models is costly and time consuming, significantly limiting their practical scalability and deployment \cite{difilippo2017system, brogan2021deep, saeed2021robust}.

Aside from perception challenges, learning robust robotic policies for unfastening task is further complicated by sensitivity to minor positional inaccuracies and force deviations. The unfastening task inherently involves sequential, contact-rich manipulation requiring sub-millimeter precision and careful regulation of forces to avoid damage to both the robot and the disassembly components. Additionally, real-world environments introduce variability due to lighting changes, surface reflections, and background clutter, all of which can degrade sensor perception and affect policy robustness. Learning stable and adaptive control under these uncertainties requires sophisticated multi-modal integration strategies—combining visual, proprioceptive, and force feedback—in a unified framework. 

\begin{figure}[t]
    \centering
    \includegraphics[width=1\linewidth, trim={0.1cm 0.1cm 0.1cm 0.1cm},clip]{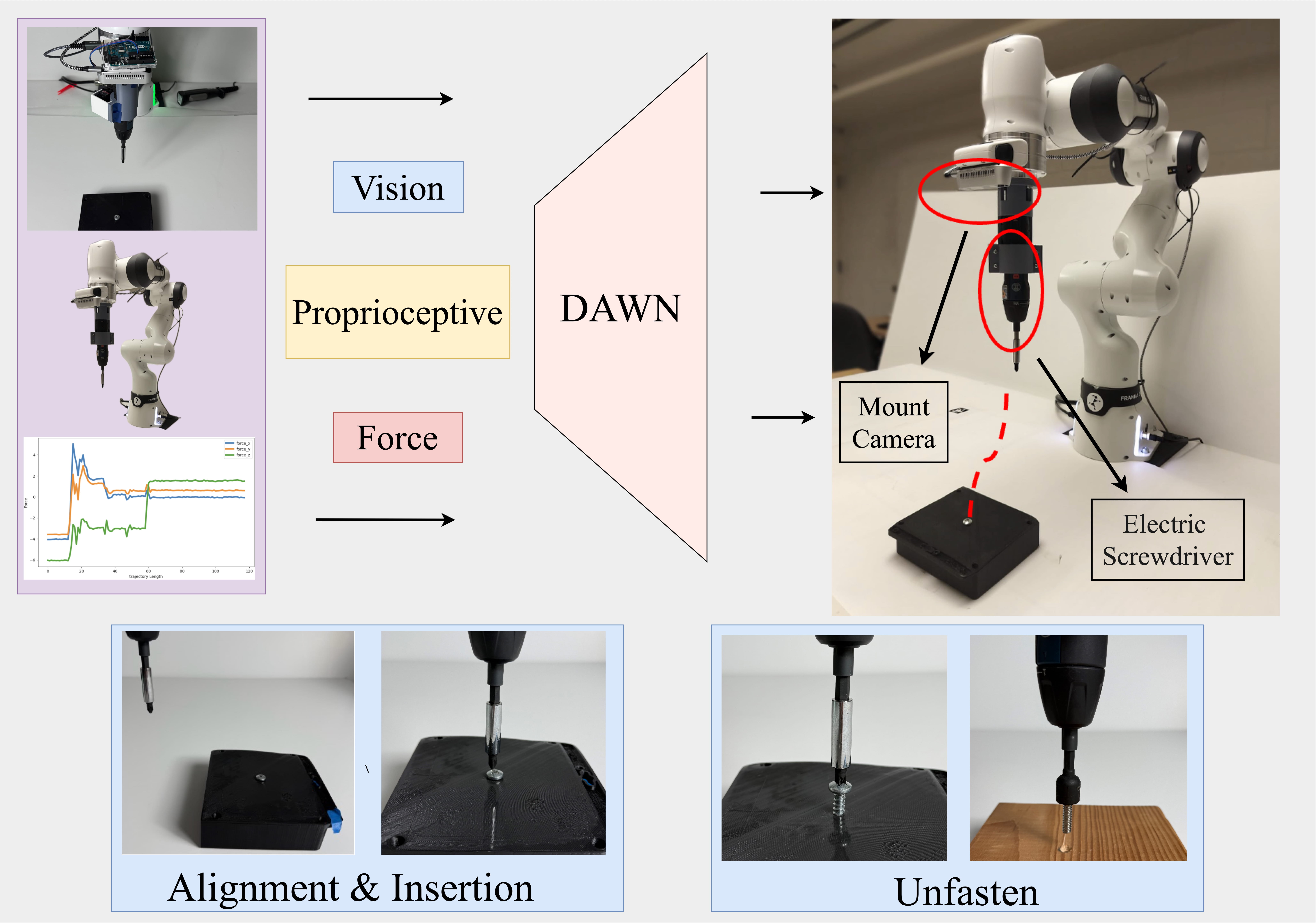}
    \caption{Overview of DAWN, a hierarchical model for solving sequential, contact-rich, long-horizon unfastening tasks.}
    \label{fig:overview-diagram}
    \vspace{-0.8cm} 
\end{figure}

To overcome these challenges, we introduce \textbf{DAWN} (Diffusion-based Alignment and unfastening Workflow Network), for learning robust and precise robotic unfastening task, see Figure \ref{fig:overview-diagram}. Drawing inspiration from human skill execution and leveraging recent advancements in diffusion-based imitation learning (IL), DAWN decomposes the complex, sequential task of unfastening task into two distinct stages: (1) alignment \& insertion, and (2) unfastening. Our proposed framework comprises three specialized components: \textbf{\textit{AlignNet}}, a policy utilizing diffusion models to achieve accurate screwdriver tip alignment and stable insertion; \textbf{\textit{LatchNet}}, a classifier that employs visual and force feedback for robust stage transition detection; and \textbf{\textit{UnfastNet}}, a policy tailored for reliably performing the unfastening operation. This structured, multi-modal approach enables dynamic policy switching, robust sensory integration, and high precision, significantly enhancing policy generalization and adaptability in real-world scenarios. Our main contributions are as follows:



(a) We present DAWN, a novel hierarchical framework designed for high-precise, contact-rich, unfastening task. To the best of our knowledge, this is the first work which addresses the robotic unfastening task with learned policies on a real-world setting. (b) In precise alignment \& insertion, DAWN achieves a success rate of $50 \% \pm 8.3\%$ comparable to expert teleoperators, with dynamic track and multi-modal sensing capabilities. Furthermore, in our ablation study, we evaluated the impact of various sensing modalities on DAWN performance. (c) We show that DAWN’s unfastening capabilities generalize well across a diverse range of fasteners, relying solely on force feedback and operating within strict safety constraints. (d) We extensively validate our policy in real-world settings using a Franka Emika robot equipped with an indigenous electric screwdriver. Our results highlight DAWN's capability to perform long-horizon sequential unfastening with a substantial success rate.






The remainder of this paper is structured as follows: Section II discusses relevant literature on robotic unfastening and IL, Section III details our proposed DAWN framework, including data collection, network architecture, and training strategies; Section IV describes experimental validation, results, and analysis; and finally, Section V concludes the paper and discusses potential future directions.


%% file: sections/litreview.tex
\section{Related Work}

\subsection{Unfastening Task}


Successfully automating unfastening necessitates precise screw localization, robust control strategies, and stable tool engagement with fasteners. Recent advances in deep learning have shifted considerable research focus toward detecting, classifying, and localizing screws using computer vision techniques \cite{difilippo2017system, brogan2021deep, saeed2021robust, mangold2022vision, li2021accurate, zhang2023novel}. Despite notable progress, vision-based methods often encounter positional inaccuracies and typically do not address the entire unfastening pipeline, from precise alignment to force-controlled unscrewing.

In parallel, various specialized tools have been proposed explicitly for robotic unfastening tasks \cite{li2020unfastening,seliger2001flexible,zuo2002novel}. However, these tools frequently struggle to achieve a stable connection with the screw head, especially when screws are tightly fastened, leading to high operational failure rates. For instance, \cite{li2020unfastening} introduces a collaborative robotic setup to sequentially unfasten hexagonal-headed screws using an electric tool but relies heavily on pre-known screw positions and precise initial tool placement. Similarly, prior work leveraging computer vision to assist unfastening tasks—primarily demonstrated on smaller electronics like laptops and hard drives—often lacks detailed analysis of positional errors, force regulation strategies, or methods to reliably handle environmental variations \cite{difilippo2017system, zhou2024towards}. Furthermore, such approaches typically require sophisticated experimental setups that complicate their adaptation and scalability to complex, and large-scale products. 

\subsection{Imitation learning}

Solving long-horizon, contact-rich manipulation tasks remains challenging for robot decision-making and learning-based methodologies \cite{haldar2024baku}. While traditional reinforcement learning (RL) has shown promise \cite{kumar2020conservative, kostrikov2021offline}, its practical deployment is hindered by the difficulty in defining suitable reward functions and the significant interaction requirements for training. IL, which enables robots to mimic expert demonstrations directly, provides a viable alternative by circumventing explicit reward definitions and reducing the demand for extensive real-world interactions \cite{zare2024survey}. However, IL approaches typically suffer from distributional shift—incremental prediction errors that compound over long task horizons—leading to brittle or suboptimal policies. Techniques such as DAgger \cite{ross2011reduction} or structured data augmentation \cite{florence2019self} have been proposed to mitigate these shifts but often introduce additional complexity or rely on assumptions unsuitable for precise, contact-rich tasks.

Recent advancements leveraging diffusion-based generative models have demonstrated considerable promise in capturing complex, multi-modal action distributions, significantly improving robustness in imitation-based policies \cite{pearce2023imitating, chi2023diffusion}. Furthermore, multi-modal sensor integration approaches, combining vision, proprioception, and force or tactile feedback, have increasingly shown advantages in addressing manipulation robustness \cite{lin2024learning, pattabiraman2024learning}. However, the effective integration of these modalities in sequential, precision-demanding tasks like unfastening remains relatively unexplored, motivating our adoption of a multi-modal diffusion-based IL framework.

%% file: sections/methodology.tex
\section{METHODOLOGY} 

\begin{figure*}[thpb]
    \includegraphics[height=0.38\linewidth]{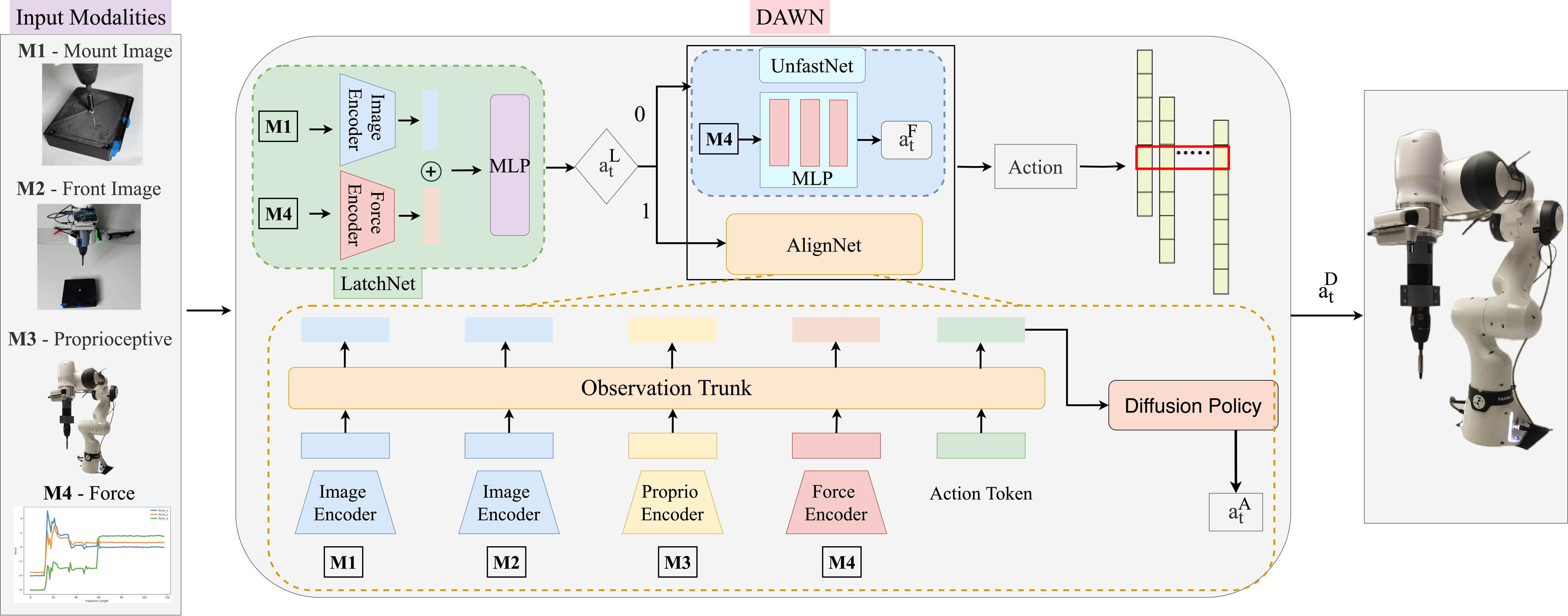}
    \caption{The architecture of the hierarchical framework, DAWN, consists of three components: (a) LatchNet, (b) AlignNet, and (c) UnfastNet. (a) LatchNet: Uses ResNet-18 to encode mount images and an MLP encoder for force input to predict stage transitions.(b) AlignNet: Uses ResNet-18 to encode mount images, an MLP encoder for force input, and proprioceptive input. An action token, along with these modalities, is passed through a transformer decoder to obtain an observation representation. This observation representation is then processed by the diffusion policy to generate the action sequence.(c) UnfastNet: Uses an MLP for force input to predict actions. }
    \label{fig:model architecture}
    \vspace{-0.6cm}
\end{figure*}

In this section, we outline our methodological approach for learning robotic unfastening task. First, we describe the procedure employed for collecting high quality expert demonstration data through teleoperation. Subsequently, we highlight the key design decisions and architectural considerations essential for effectively training and deploying the proposed DAWN framework.

\subsection{Teleoperation, Data Collection and Pre-processing }

Real-world human demonstration datasets have been collected via teleoperating the Franka Emika 7 DOF robot arm {with inbuilt force/torque sensor at end-effector}. An X-box controller is used to perform the 3-dimensional positional control of the manipulator. The orientation of the end-effector is constrained to remain vertically downward throughout the task, see Figure \ref{fig:model architecture}. We collected multi-modal experimental data at 10 Hz. The datasets are composed of the proprioceptive states of the manipulator, RGB images captured by three Intel RealSense D435i (mounted on the manipulator, front, and right sides), and action commands. 



\textit{Vision}: At each time step, we obtain three RGB images at a resolution of 240×320. 

\textit{Proprioceptive States}: The data include the end-effector pose, joint positions, and screwdriver tip pose. 


\textit{Force Feedback}: We obtain continuous force data from the built-in force/torque sensor. 

\textit{Output control}: The output control action is acquired, which corresponds to the desired pose of the end-effector. 

\textit{{Data Normalization}}: All collected sensory modalities were normalized to the range [-1,1] before feeding them to subsequent encoder networks \cite{chi2023diffusion}.



\subsection{Model Architecture}

The objective of our proposed method is to learn unfastening policy, DAWN, $\pi_{\theta}^{D} : \{o_t\} \rightarrow \{a_t\}$, directly from offline demonstrations. The policy maps observations $o_t$  from an observation space to continuous actions $a_t$. Our proposed architecture, illustrated in Figure \ref{fig:model architecture}, composed of three main components: (a) AlignNet: $\pi_{\theta}^{A} : \{o_t^A\} \rightarrow \{a_t^A\}$, (b) LatchNet: $\pi_{\theta}^{L} : \{o_t^L\} \rightarrow \{a_t^L\}$,  and (c) UnfastNet: $\pi_{\theta}^{F} : \{o_t^F\} \rightarrow \{a_t^F\}$. Each component is trained separately.  

Similar to \cite{chi2023diffusion},  we condition each component policy on a historical sequence of observations $o_t = [\textbf{\textit{o}}_{t - T_o}, \, \ldots, \, \textbf{\textit{o}}_t]$, where $T_0$ denotes the number of observations, and each observation $o_t$ is of dimension $|o|$. The policy maps the $o_t $  to a horizon of future actions $a_t = [\textbf{\textit{a}}_t, \, \ldots, \, \textbf{\textit{a}}_{t + T_p}]$, where $T_p$ denotes the prediction horizon. To enhance motion smoothness and mitigate the covariate shift inherent in low-data imitation learning regimes, we employ exponential temporal ensembling and action chunking strategies \cite{zhao2023learning}. Consequently, at each execution step on the real robot, we execute only a subset of predicted actions $a_t = [\textbf{\textit{a}}_t, \, \ldots, \, \textbf{\textit{a}}_{t + T_a}]$, where $T_a$ represents the executed action horizon,  $T_a < T_p$. {For alignment \& insertion stage, the policy predicts absolute end-effector positions, whereas for unfastening it predicts delta actions, i.e., changes relative to the current end-effector position, reflecting the distinct precision and compliance requirements of each stage}. 



\vspace{0.20 cm}

\textbf{a. AlignNet}
\vspace{0.10 cm}

To learn our alignment \& insertion policy $\pi_{\theta}^{A}(a_t^A | o_t^A)$ from expert demonstrations $\mathcal{D^A} = \{\tau_i\}_{i=1}^{N}
    $, where each trajectory $\tau_i = \{(o_i^0, a_i^0), \dots, (o_i^T, a_i^T)\}$ consists of observation-action pairs, we adopt a transformer decoder network to learn a rich, multi-modal observation representation \cite{haldar2024baku}. Leveraging this combined representation, a diffusion-based policy \cite{chi2023diffusion} generates the action sequence, as illustrated in Figure \ref{fig:model architecture}b. Ablation analysis demonstrates that multi-modal integration is critical for high-precision manipulation tasks (see Table I). AlignNet consists of three main components: (i) Encoder, (ii) Observation Trunk, and (iii) Diffusion model. 



\vspace{0.20 cm}
(i) \underline{Encoders}: \textit{Vision}: Following the prior work on visuomotor diffusion policy \cite{chi2023diffusion}, we employ a ResNet-18 \cite{he2016deep} network for visual encoding and all the BatchNorm \cite{ioffe2015batch} layers in the network are replaced with GroupNorm \cite{wu2018group}. The vision encoder outputs embeddings $(e_t^v)$ of dimension $|e_t^v| = T_o \times 256$. 


{\textit{Proprioception}: We only used end-effector position ($x$,$y$,$z$). The input is processed through a 2-layer perceptron with dropout (0.3), ReLU activations, and 256 hidden units, producing embeddings $(e_t^p)$ of dimension $|e_t^p| = T_o \times 256$.}



\textit{Force}: Similarly, force sensor data is encoded using an identical 2-layer perceptron, producing output embeddings $(e_t^f)$ of dimension $|e_t^f| = T_o \times 256$.


\vspace{0.10 cm}

(ii) \underline{Observation Trunk}: Modality-specific embeddings $(e_t^p, e_t^v, e_t^f)$ are encoded into tokens and fed into the transformer decoder. Additionally, a learnable action token $\{e_t^a\}$ is included, which interacts with other tokens to generate a comprehensive and contextually rich observation representation $\{o_t^a\}$ of dimension $|o_t^a| = 256$.


\vspace{0.10 cm}
(iii) \underline{Diffusion Model}: We adopted 1D temporal CNN-based diffusion network $\varepsilon_{\theta}$. Similar to the \cite{chi2023diffusion}, we use DDPM \cite{ho2020denoising} noise scheduler with $k = 100$ for training and $k = 30$ for evaluation. The diffusion model predicts an action sequence of length $T_p = 16$ with an observation horizon of $T_o = 2$ and an action horizon of $T_a = 8$.

\vspace{0.10 cm}

\textbf{b. LatchNet}
\vspace{0.10 cm}

The LatchNet (Figure \ref{fig:model architecture}a) policy $\pi_{\theta}^{L}(o_t^L)$ leverages visual and force feedback modalities to classify the transition between alignment \& insertion and unfastening stages. Specifically, it predicts whether the screwdriver tip is accurately and securely latched onto the screw head, indicating readiness for the subsequent unfastening procedure. To capture rich semantic features from visual observations, we use a ResNet-18 network pretrained on ImageNet, keeping its weights frozen during training. The vision encoder outputs embeddings $(e_t^v)$ of size $|e_t^v| = T_o \times 256$. Concurrently, force sensor data is processed through a two-layer perceptron, producing embeddings $e_t^f$ of dimension $|e_t^f| = T_o \times 256$, with $T_o =2$. The embeddings visual and force encoder embeddings concatenated and subsequently passed through a binary classifier implemented as a two-layer perceptron. The network is trained using a binary cross-entropy loss function: 
\begin{equation}
    \mathcal{L}^L = - \left[ a_t^L \log({\hat a_t^L}) + (1 - a_t^L) \log(1 - {\hat a_t^L}) \right]
\end{equation}

where $a_t^L \in \{0, 1\}$ is the ground truth label, and $\hat a_t^L \in \{0, 1\}$ is the model's prediction.



\vspace{0.10 cm}

\textbf{c. UnfastNet}
\vspace{0.10 cm}



To learn the UnfastNet (Figure \ref{fig:model architecture}a) policy $\pi_{\theta}^{F}(a_t^F|o_t^F)$, we collected expert demonstration trajectories $\mathcal{D}^{F} = \{\tau_j\}_{j=1}^{M}$, with each trajectory consisting of observation-action pairs, represented as $\tau_j = \{(o_j^0, a_j^0), \dots, (o_j^T, a_j^T)\}$. UnfastNet is implemented as a lightweight three-layer perceptron featuring dropout (0.3), ReLU activation functions, and 128 neurons per hidden layer. The model outputs predicted action sequences with dimension $T_a \times |a_t^F|$, where $T_a$ denotes the executed action horizon, and $|a_t^F|$ indicates the dimensionality of the continuous action commands.

The UnfastNet policy is trained using a BC loss:
\begin{equation}
    \mathcal{L}^{F} = \mathbb{E}_{(o_t^F,a_t^F) \sim \mathcal{D}^{F}}\|a_t^F - \pi_{\theta}^{F}(o_t^F)\|^2.
\end{equation}

During policy execution, UnfastNet explicitly satisfies two critical operational constraints: (i) maintaining stable contact between the screwdriver tip and screw head, and (ii) limiting reaction forces at the robot’s end-effector to remain consistently below 20~N, thereby ensuring mechanical safety and robust task execution.

%% file: sections/experiments.tex
\section{Experiments}

For all experiments, we use a Franka Emika Panda robot arm equipped with a custom-designed electric screwdriver mounted on the end-effector, see Figure \ref{fig:model architecture}. All models are trained and evaluated on a workstation with an NVIDIA RTX 4080 GPU. We did an exhaustive evaluation of the proposed DAWN to validate its performance, efficiency, and generalization capabilities by demonstrating its ability to successfully complete unfastening task in a real-world setting (an average of \textbf{1000} experiments in total). The majority of our experimental analysis focuses extensively on the alignment \& insertion stage, as it represents the critical bottleneck in automated unfastening and directly impacts overall success rate. Even a slight misalignment in this stage inevitably leads to downstream task failure, making it essential for robust policy development and evaluation.  

For evaluation, multiple checkpoints are saved during training. To assess performance, we conduct  10 real-world trials per checkpoint for different screw head positions as shown in Figure \ref{fig: Figure 4}. Each policy is trained across three random seeds, and the average success rate is reported. The learnt policies are deployed at a frequency of 10 Hz {\cite{haldar2024baku, chi2023diffusion}}. We designed experiments to address the following key questions:

\begin{itemize}
    \item [(a)] How effectively does DAWN learn each sub-task (alignment \& insertion, stage classification, unfastening)?
    \item [(b)] What impact do different sensory modalities have on AlignNet, LatchNet, and UnfastNet?
    \item [(c)] How robustly does DAWN generalize to unseen task variations?
\end{itemize}



Further, we conducted ablation studies to evaluate (i) the influence of the number of demonstrations, and (ii) the role of the transformer decoder in DAWN's performance. Finally, we report DAWN’s performance on the complete unfastening task.



 \begin{table*}
    \centering
    \caption{ \normalfont  Success rates (out of 10) averaged over three seeds for policies trained with different input modalities.} 
    \begin{tabular}{c|c|c|c|c|c|c|c}
    \toprule
    \multicolumn{4}{c}{Input Modalities} & \multicolumn{4}{c}{AlignNet} \\
    Mount Camera & Robot Proprio & Force & Front Camera & Alignment & Insertion & Proximity & Success Rate\\
    \midrule
    Yes  & Yes  &  No  & No  & $5.0 \pm 0.8$ & $8.3 \pm 1.2$ & $8.3 \pm 1.2$  & $5.0 \pm 0.8$\ \\
    Yes  & No  & No  & No   & $1.4 \pm 1.2$ & $0.0 \pm 0.0$ & $2.7 \pm 2.3$ & $0.0 \pm 0.0$ \\
    Yes & Yes & Yes & No  & $0.7 \pm 0.4$ & $0.7 \pm 0.4$ & $4.0 \pm 0.8$  & $0.3 \pm 0.4$ \\
    Yes & Yes & No & Yes  & $1.0\pm 0.0$ & $0.0 \pm 0.0$ & $3.5 \pm 0.5$ & $0.0 \pm 0.0$\\
    \bottomrule
    \end{tabular}
    \label{Table 1}
    
\end{table*}

\begin{figure}
    \includegraphics[width=1\linewidth]{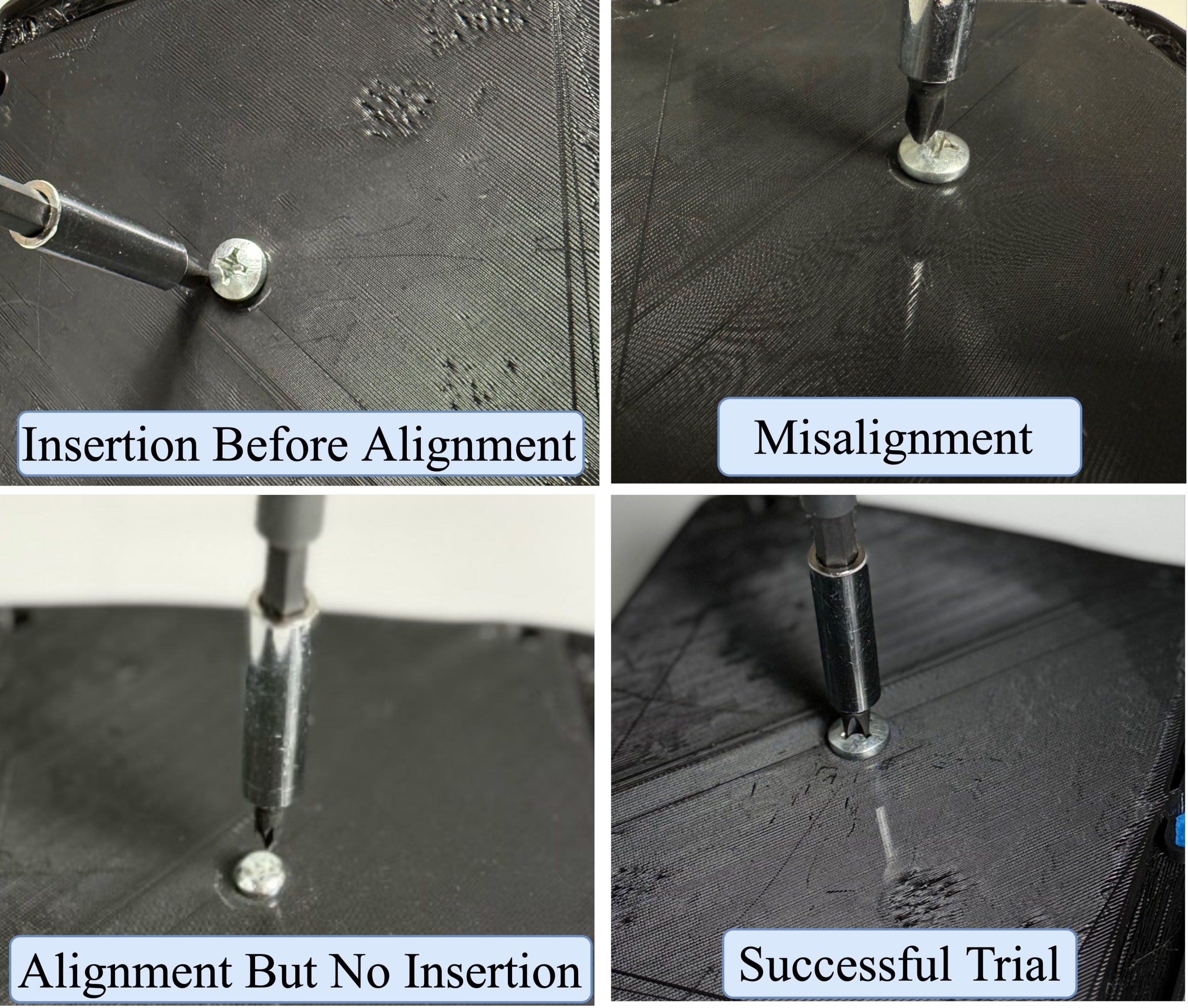}
    \caption{Common reasons behind unsuccessful policy roll-outs.}
    \label{fig: Figure 3}
    \vspace{-0.6cm} 
\end{figure}

\begin{table*}
    \centering
    \caption{\normalfont We compare the best DAWN policy’s performance against three human teleoperators (two experts and one non-expert) and a baseline diffusion policy \cite{chi2023diffusion}. We also highlight the importance of vision feedback in these precise, contact-rich robotic tasks.}
    \begin{tabular}{c|ccc|c|ccc} 
    \toprule
     & \multicolumn{3}{c|}{Human Operator} & \multicolumn{1}{c|}{Diffusion Policy} & \multicolumn{3}{c}{ResNet-18} \\
     & Exp 1 & Exp 2 & Non-Exp 1 & W/O Obs-Trunk & Scratch & Finetuned & $4^{th}$-Layer Finetuned\\
    \midrule
    Alignment & $6.0 \pm 0.0$ & $6.0 \pm 0.0$ & $3.0 \pm 0.0$ & $2.0 \pm 0.0$ & $5.0 \pm 0.8$ &$0.0 \pm 0.0$ & $0.0 \pm 0.0$ \\
     Insertion & $10.0 \pm 0.0$ & $10.0 \pm 0.0$ & $10.0 \pm 0.0$ &  $0.0 \pm 0.0$ & $8.3 \pm 1.2$ & $0.0 \pm 0.0$& $0.0 \pm 0.0$ \\
     Proximity & $10.0 \pm 0.0$ & $10.0 \pm 0.0$ & $10.0 \pm 0.0$ &  $2.0 \pm 0.0$ & $8.3 \pm 1.2$   &$2.0 \pm 0.0$& $10.0 \pm 0.0$\\
     Success Rate & $6.0 \pm 0.0$ & $6.0 \pm 0.0$ & $3.0 \pm 0.0$ &  $0.0 \pm 0.0$ & $5.0 \pm 0.8$\ &$0.0 \pm 0.0$& $0.0 \pm 0.0$\\
    Avg. Time (sec) & $23.6 \pm 8.2$ & $31.75 \pm 6.1 $& $23.9 \pm 5.6$ &  - &  $30.2 \pm 4.8$ & - & -\\
    \bottomrule
    \end{tabular}
    \label{tab: Table 2}
    
\end{table*}

\begin{table*}[t]
    \centering
    \begin{minipage}[t]{0.48\linewidth}
        \centering
        \caption{\normalfont Performance of DAWN policy on the number of demonstrations}
        \begin{tabular}{c|c|c|c|c}
            \toprule
            Number of Demos & Alignment & Insertion & Proximity & Success \\
            \midrule
            101 & $2.0 \pm 0.8$ & $2.7 \pm 1.7$ & $5.0 \pm 2.9$ & $0.4 \pm 0.4$ \\
            151 & $2.0 \pm 1.4$ & $2.0 \pm 1.4$ & $3.7 \pm 1.2$ & $0.7 \pm 0.4$ \\
            199 & $5.0 \pm 0.8$ & $8.3 \pm 1.2$ & $8.3 \pm 1.2$ & $5.0 \pm 0.8$ \\
            \bottomrule
        \end{tabular}
        \label{tab: Table 3}
    \end{minipage}
    \hfill 
    \begin{minipage}[t]{0.48\linewidth}
        \centering
        \caption{\normalfont Effect of object variations on AlignNet performance}
        \begin{tabular}{c|c|c|c|c}
            \toprule
            Variation & Alignment & Insertion & Proximity & Success \\
            \midrule
            Size     & $2.0 \pm 0.0$ & $6.0 \pm 1.0$ & $6.0 \pm 1.0$ & $0.0 \pm 0.0$ \\
            Type     & $1.0 \pm 0.0$ & $0.0 \pm 0.0$ & $7.0 \pm 0.0$ & $0.0 \pm 0.0$ \\
            Material & $0.0 \pm 0.0$ & $4.0 \pm 0.0$ & $6.0 \pm 0.0$ & $0.0 \pm 0.0$ \\
            \bottomrule
        \end{tabular}
        \label{tab: table 4}
        
        
    \end{minipage}
\end{table*}


\subsection{Task Description}


{Each stage of the unfastening task presents distinct challenges and requires specific evaluation criteria, described below.} 

(i) Alignment \& Insertion: This stage involves precise engagement of the electric screwdriver mounted on the robot’s end-effector with the Phillips screw head located on a box placed randomly within a workspace of 40 cm $\times$ 15 cm (Figure \ref{fig: Figure 4}). We collected 199 expert demonstrations to train this stage. To clearly evaluate task execution beyond a binary success metric, we define explicit terms to provide precise criteria to evaluate distinct aspects of the alignment \& insertion stage, facilitating better analysis of failure modes, and improved reproducibility as seen in Figure 3. These terms are defined as:

\begin{itemize}
    \item Alignment: Precise vertical alignment of the screwdriver tip directly above the central axis of the screw head.
    \item Proximity: The screwdriver tip is within a 2 cm radius from the screw head's central axis.
    \item Insertion: Downward movement initiated once the screwdriver tip is within the defined proximity radius.
    \item Success: Stable and secure latching of the screwdriver tip with the screw head.
\end{itemize}

(ii) Latching: 
Achieving a stable latch is crucial, as misalignment or insufficient contact can lead to slippage, increased reaction forces which can damage robot, or task failure. For robust LatchNet training, we created a diverse dataset of 6,000 mount-camera images capturing various screwdriver-to-screw interactions and associated force readings indicative of contact stability. Extensive data augmentation, including random flipping, lighting variations, color jittering, rotational transformations, masking for images, and Gaussian noise for force measurements, was applied to improve model robustness and generalization to real-world conditions.


(iii) Unfastening: The unfastening demands precise force and motion control. We collected 15 demonstrations where the electric screwdriver unfasten a Phillip screw. For these demonstrations, we start when the screwdriver and screw head is properly engaged. During the unfastening, we aimed to maintain continuous axial motion and reaction force. Several prominent failure modes were observed, including excessive reaction force experienced by the robot, which can damage the robot, break the screw or compromise the integrity of the object. To ensure successful, safe and smooth unfastening, the task demands satisfying the two operational constraints. 



\vspace{-0.05cm}
\subsection{Capability of DAWN to Learn}

\subsubsection{AlignNet}
The alignment \& insertion stage demands precise control to engage the screwdriver tip accurately with the screw head, as illustrated in Figure \ref{fig: Figure 3}. Unlike typical plug insertion tasks previously explored \cite{pattabiraman2024learning}, which generally involve surface contact followed by sliding motions to locate the cavity, our setup intentionally positions the screw head above the surface level. This design requires the policy to achieve accurate vertical alignment before initiating insertion, eliminating reliance on sliding motions. Specifically, the screwdriver tip must be precisely aligned within a 4 mm margin directly above the screw head's central axis; deviations beyond this margin result in task failure (Figure \ref{fig: Figure 3}). This small margin of error in alignment constraint significantly increases task complexity.

To evaluate the policy's capability and task difficulty, we benchmarked AlignNet against human teleoperation performance and diffusion policy \cite{chi2023diffusion}. {Three human operators, two with approximately 10–15 minutes of training and one with no prior experience—performed 10 trials each under varying target screw positions using an X-Box controller for teleoperation.} Notably, AlignNet's success rates closely matched expert human performance, as presented in Table II. While the non-trained quickly improved over repeated trials, their overall success rate remained lower, reinforcing the inherent challenge and precision demands of the task. These findings validate AlignNet's effectiveness and demonstrate its ability to match human-level teleoperation performance in executing alignment \& insertion. {Our results show that training the vision encoder from scratch outperforms both fully fine-tuned and only final layer fine-tuned ResNet-18, as reported in Table \ref{tab: Table 2}}.

\subsubsection{LatchNet}
LatchNet is trained using a binary cross-entropy loss with the Adam optimizer (learning rate $= 10^{-4}$, weight decay $= 10^{-6}$, batch size $= 256$). Evaluation metrics include accuracy, precision, recall, and F1 score. LatchNet achieves high performance across these metrics—0.90 accuracy, 0.90 precision, 0.98 recall, and 0.94 F1—confirming its robustness in classifying stable latching. The notably high recall value underscores LatchNet's effectiveness in minimizing false negatives, crucial for practical deployment in real-world robotic tasks where reliable stage transitions are essential to prevent task failure and potential robot damage.



\subsubsection{UnfastNet}
While diffusion-based policies performed well for alignment \& insertion, their latency in trajectory generation caused motion mismatches during unfastening, often triggering safety stops due to excessive force.

To overcome these limitations, we employed a simple lightweight network capable of predicting short-horizon actions directly. However, this approach exhibited two primary failure modes: premature screwdriver disengagement due to low initial reaction forces, and force threshold breaches when initial forces were excessively high. Addressing these challenges, we incorporated heuristic-based force regulation and continuous upward motion constraints, significantly enhancing task reliability. {To address this, we incorporated heuristic-based force regulation by turning off the screwdriver when the force feedback reached $\geq 10\text{N}$, which significantly enhanced task reliability.}

\subsection{Impact of Modalities on Performance}

In this section, we systematically evaluated the influence of various input modalities on each component within the DAWN hierarchical framework.


\subsubsection{AlignNet}
For this stage, we evaluated multiple modality configurations involving mount camera images, front camera images, force feedback, and proprioceptive data, Table I. Our findings indicate that the combination of mount camera images and proprioceptive information alone is sufficient to achieve high alignment success rates. Incorporating additional camera inputs increased computational complexity and training duration but resulted in no notable performance gain, and in some cases, led to reduced accuracy. Similarly, including force feedback did not improve alignment performance, reflecting the specific nature of this task, which fundamentally requires precise spatial alignment rather than force-sensitive adjustments.

\subsubsection{LatchNet}
The effectiveness and robustness of LatchNet significantly depend on input modalities used for accurately classifying stable engagements and subsequent stage transitions. Analogous to human perception, LatchNet integrates visual and force feedback modalities to reliably determine screwdriver latching conditions. Evaluations indicate that employing either modality in isolation introduces specific vulnerabilities: vision-only classifiers struggle with screw occlusions in camera views, causing erroneous predictions, while force-only classifiers may incorrectly signal latch engagement due to low reaction forces or contact with unintended surfaces. Consequently, combining both visual and force inputs is crucial for accurately and reliably detecting stable latch.



\subsubsection{UnfastNet}

\begin{figure*}
    \centering
    \includegraphics[width=\linewidth]{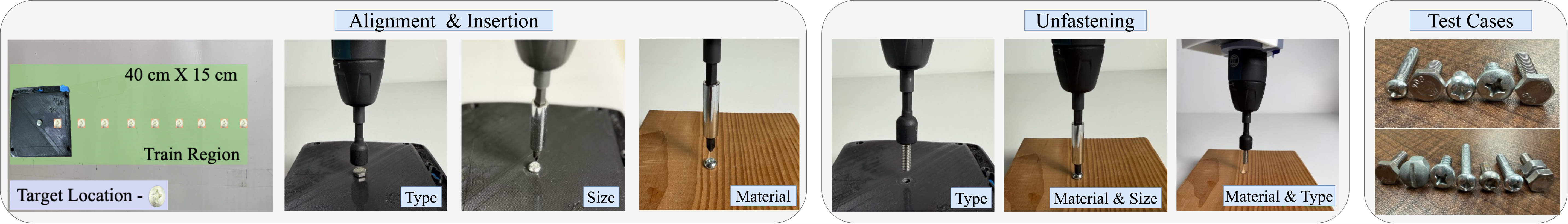}
    \caption{Overhead view illustrating the different target object positions used for training and evaluation. The light green box indicates the extent of positional variation in the training data, and the evaluation locations are labeled. Different types and sizes of fasteners, and materials were employed for assessing DAWN's performance.}
    \label{fig: Figure 4}
    \vspace{-0.6cm}
\end{figure*}

Utilizing exclusively force-based input significantly reduces latency between action prediction \& execution,  ensure environmental invariance and maintain consistent, scalable performance. Evaluations confirmed that UnfastNet with heuristics maintains a robust perfect success rate across various screw types, dimensions, colors, and materials (Figure \ref{fig: Figure 4}, Table \ref{tab:screw_variants_horizontal}). {Notably, all experiments were conducted with fastener heads oriented flat (parallel to the surface). Minor misalignments  between the screwdriver and fastener head during unfastening were compensated by the compliance of the screwdriver bit. Unfastening at non-parallel angles was not evaluated in this work but can be addressed through appropriate coordinate transformations.}

\subsection{Generalization and Robustness in Complex Environments and Unseen Task Variations}

We rigorously evaluated DAWN's generalization capabilities and robustness by testing the learned policies on previously unseen task variations. 


\subsubsection{Alignment \& Insertion}

The results in Table IV highlight specific challenges for AlignNet when generalizing to variations in screw characteristics such as size, type, and material. We observed that when the screw size was smaller than the training data, AlignNet consistently managed to position the screwdriver tip within a 2 cm proximity radius. However, due to the reduced size, precise insertion attempts consistently failed. Additionally, tests involving hexagonal bolts revealed significant difficulties; although visually identifiable by the model, successful task execution was hindered because the screwdriver tip fully occluded the bolt, disrupting visual feedback. Furthermore, when evaluating screws placed in different materials, such as wood, reduced visual contrast between the screw and its background substantially impaired the alignment accuracy, resulting in no successful trials. These detailed observations suggest the current visual encoding and training procedures are overly specialized, limiting generalization. To enhance robustness and adaptability, future improvements should incorporate more diverse training data, advanced data augmentation techniques, and adaptive visual encoding strategies.

\subsubsection{Stage Transition}

For robust and reliable stage transitions in diverse and previously unseen scenarios, LatchNet's generalization capabilities are essential. To thoroughly evaluate its robustness, we systematically introduced variations in experimental conditions, including changes in the fastener type, size, and fluctuations in force measurements. While LatchNet consistently demonstrated accurate predictions of stage transitions across most conditions, we identified practical limitations in scenarios involving bolts, where the screwdriver tip completely covers the bolt head during unfastening, rendering visual inputs ineffective. This occlusion poses significant challenges for the vision-based component of LatchNet, occasionally resulting in misclassifications. These observations highlight the importance of enhancing multi-modal integration or incorporating alternative sensing methods to reliably manage visually obscured scenarios, ensuring stable and dependable task execution.



\subsubsection{Unfastening}


We assessed UnfastNet's generalization capability on a variety of previously unseen fasteners, including screws and bolts of different pitches, lengths, and sizes, as depicted in Figure \ref{fig: Figure 4}. {UnfastNet consistently executed unfastening tasks with high reliability across all tested fastener types. For brevity, only a subset of results is reported in Table \ref{tab:screw_variants_horizontal}, all of which demonstrate perfect success.} We observe certain challenges emerged specifically with bolts, where the initial reaction force was occasionally insufficient; despite this, UnfastNet was still able to perform successful unfastening attempts. This particular issue highlights the need for improved initial force detection and adaptive strategies to enhance reliability in scenarios involving bolts. These findings demonstrate that UnfastNet can effectively perform robust force and motion control in real-world settings.

\begin{table}[h]
\centering
\caption{\normalfont  Unfastening success rates across different fastener types.}
\begin{tabular}{|c|c|c|c|c|}
\hline
\textbf{Type} & {Phillips} & {Security Torx} & Slotted &  {Hex } \\
\hline
\textbf{Success } & 5/5 & 5/5 & 5/5 & 5/5  \\
\hline
\end{tabular}
\label{tab:screw_variants_horizontal}
\end{table}
\vspace{-0.5cm}

\subsection{Ablation Study}


\subsubsection{Learning Efficiency}
We empirically investigated how varying the number of demonstrations impacts the learned policy's real-world performance, specifically focusing on the alignment \& insertion stage, see Table III.
As mentioned in offline learning, traditional evaluation metrics such as validation loss or MSE are not reliable indicators of real-world performance \cite{mandlekar2021matters, chi2023diffusion}. Instead, the learned policy must be deployed on the physical robot for proper evaluation. As illustrated in Table III, increasing the demonstration count clearly improved success rates, indicating that additional training data directly contributes to enhanced robustness and task execution reliability. This observation highlights the potential benefit of larger demonstration datasets to further improve real-world policy effectiveness.


\subsubsection{Role of the Observation Trunk}
In scenarios such as precise screwdriver engagement, heavy reliance on visual inputs by the diffusion policy \cite{chi2023diffusion} leads to frequent insertion failures due to insufficient positional awareness. The observation trunk addresses this limitation by generating enriched observation representations, effectively balancing visual and positional information. Experimental evaluations (see Table II) demonstrate that policies incorporating the observation trunk significantly outperform those without, particularly improving insertion success rates by ensuring appropriate weighting of positional information.

\subsection{Complete Unfastening Task}

We evaluated DAWN on the complete unfastening task. For each stage, the best-performing model configuration was selected and deployed across 10 real-world trials under consistent target setups. The system achieved a success rate of $5.0 \pm 0.8$\%. Our results show that overall task success is critically dependent on the accuracy of the alignment and insertion stage. In particular, when the screwdriver tip is perfectly latched onto the screw head, unfastening consistently achieves a perfect success rate. These findings highlight the pivotal role of precise alignment and insertion in ensuring robust and reliable execution of the full unfastening process.

%% file: sections/conclusion.tex
\section{Conclusion}


In this study, we introduced DAWN, a hierarchical robotic framework designed for robust, autonomous execution of precise, contact-rich unfastening tasks. 
To simplify learning of this complex task, we subdivided unfastening task into sequential stages: (i) alignment \& insertion, and (ii) unfastening. DAWN achieves alignment \& insertion accuracy comparable to trained human teleoperators, highlighting the effectiveness of carefully integrated visual and proprioceptive information coupled with an observation trunk. For the unfastening stage, DAWN demonstrates robust force and motion control, consistently achieving perfect success rate under stringent force constraints.


Despite these strengths, several limitations remain. The alignment \& insertion stage exhibits sensitivity to variations in visual conditions, particularly when encountering screw characteristics significantly differing from training conditions. Additionally, the reliance on precisely defined proximity and alignment criteria highlights the potential brittleness of the model in highly variable environments. Another critical limitation arises when dealing with bolts, as the screwdriver tip fully covers the bolt head during unfastening, rendering visual data ineffective for latching verification.




%% file: root.bbl
\begin{thebibliography}{10}

\bibitem{wang2023energy}
W.~Wang, G.~Tian, G.~Yuan, and D.~T. Pham, ``Energy-time tradeoffs for remanufacturing system scheduling using an invasive weed optimization algorithm,'' {\em Journal of Intelligent Manufacturing}, vol.~34, no.~3, pp.~1065--1083, 2023.

\bibitem{davis2019strong}
J.-M. Davis and Y.~Garb, ``A strong spatial association between e-waste burn sites and childhood lymphoma in the west bank, palestine,'' {\em International Journal of Cancer}, vol.~144, no.~3, pp.~470--475, 2019.

\bibitem{yildiz2019dcnn}
E.~Yildiz and F.~W{\"o}rg{\"o}tter, ``Dcnn-based screw detection for automated disassembly processes,'' in {\em 2019 15th international conference on signal-image technology \& internet-based systems (SITIS)}, pp.~187--192, IEEE, 2019.

\bibitem{zhang2023novel}
H.~Zhang, Y.~Zhang, Z.~Wang, S.~Zhang, H.~Li, and M.~Chen, ``A novel knowledge-driven flexible human--robot hybrid disassembly line and its key technologies for electric vehicle batteries,'' {\em Journal of Manufacturing Systems}, vol.~68, pp.~338--353, 2023.

\bibitem{zhou2022design}
C.~Zhou, B.~Engelen, I.~Zaplana, and J.~Peeters, ``Design of a robotic system for battery dismantling from tablets,'' {\em Procedia CIRP}, vol.~105, pp.~273--277, 2022.

\bibitem{zhou2023you}
C.~Zhou, W.~Sterkens, D.~J. D{\'\i}az-Romero, I.~Zaplana, and J.~Peeters, ``You only demanufacture once (yodo): Weee retrieval using unsupervised learning,'' {\em Resources, Conservation and Recycling}, vol.~190, p.~106826, 2023.

\bibitem{liu2020collaborative}
J.~Liu, Z.~Zhou, D.~T. Pham, W.~Xu, C.~Ji, and Q.~Liu, ``Collaborative optimization of robotic disassembly sequence planning and robotic disassembly line balancing problem using improved discrete bees algorithm in remanufacturing,'' {\em Robotics and Computer-Integrated Manufacturing}, vol.~61, p.~101829, 2020.

\bibitem{peeters2016tool}
J.~R. Peeters, P.~Vanegas, C.~Mouton, W.~Dewulf, and J.~R. Duflou, ``Tool design for electronic product dismantling,'' {\em Procedia CIRP}, vol.~48, pp.~466--471, 2016.

\bibitem{foo2021screw}
G.~Foo, S.~Kara, and M.~Pagnucco, ``Screw detection for disassembly of electronic waste using reasoning and re-training of a deep learning model,'' {\em Procedia CIRP}, vol.~98, pp.~666--671, 2021.

\bibitem{zhang2023automatic}
X.~Zhang, K.~Eltouny, X.~Liang, and S.~Behdad, ``Automatic screw detection and tool recommendation system for robotic disassembly,'' {\em Journal of Manufacturing Science and Engineering}, vol.~145, no.~3, p.~031008, 2023.

\bibitem{al2022machine}
A.~Al~Assadi, D.~Holtz, F.~N{\"a}gele, C.~Nitsche, W.~Kraus, and M.~F. Huber, ``Machine learning based screw drive state detection for unfastening screw connections,'' {\em Journal of Manufacturing Systems}, vol.~65, pp.~19--32, 2022.

\bibitem{klas2021kit}
C.~Klas, F.~Hundhausen, J.~Gao, C.~R. Dreher, S.~Reither, Y.~Zhou, and T.~Asfour, ``The kit gripper: A multi-functional gripper for disassembly tasks,'' in {\em 2021 IEEE International Conference on Robotics and Automation (ICRA)}, pp.~715--721, IEEE, 2021.

\bibitem{deng2022tiny}
Z.~Deng, D.~Guan, Y.~Qiao, and C.~Zhuang, ``Tiny screw and screw hole detection for automated maintenance processes,'' in {\em 2022 IEEE International Conference on Mechatronics and Automation (ICMA)}, pp.~847--851, IEEE, 2022.

\bibitem{zhou2024towards}
C.~Zhou, Y.~Wu, W.~Sterkens, M.~Piessens, P.~Vandewalle, and J.~R. Peeters, ``Towards robotic disassembly: A comparison of coarse-to-fine and multimodal fusion screw detection methods,'' {\em Journal of Manufacturing Systems}, vol.~74, pp.~633--646, 2024.

\bibitem{difilippo2017system}
N.~M. DiFilippo and M.~K. Jouaneh, ``A system combining force and vision sensing for automated screw removal on laptops,'' {\em IEEE Transactions on Automation science and engineering}, vol.~15, no.~2, pp.~887--895, 2017.

\bibitem{brogan2021deep}
D.~P. Brogan, N.~M. DiFilippo, and M.~K. Jouaneh, ``Deep learning computer vision for robotic disassembly and servicing applications,'' {\em Array}, vol.~12, p.~100094, 2021.

\bibitem{saeed2021robust}
F.~Saeed, M.~J. Ahmed, M.~J. Gul, K.~J. Hong, A.~Paul, and M.~S. Kavitha, ``A robust approach for industrial small-object detection using an improved faster regional convolutional neural network,'' {\em Scientific reports}, vol.~11, no.~1, p.~23390, 2021.

\bibitem{mangold2022vision}
S.~Mangold, C.~Steiner, M.~Friedmann, and J.~Fleischer, ``Vision-based screw head detection for automated disassembly for remanufacturing,'' {\em Procedia CIRP}, vol.~105, pp.~1--6, 2022.

\bibitem{li2021accurate}
X.~Li, M.~Li, Y.~Wu, D.~Zhou, T.~Liu, F.~Hao, J.~Yue, and Q.~Ma, ``Accurate screw detection method based on faster r-cnn and rotation edge similarity for automatic screw disassembly,'' {\em International Journal of Computer Integrated Manufacturing}, vol.~34, no.~11, pp.~1177--1195, 2021.

\bibitem{li2020unfastening}
R.~Li, D.~T. Pham, J.~Huang, Y.~Tan, M.~Qu, Y.~Wang, M.~Kerin, K.~Jiang, S.~Su, C.~Ji, {\em et~al.}, ``Unfastening of hexagonal headed screws by a collaborative robot,'' {\em IEEE Transactions on Automation Science and Engineering}, vol.~17, no.~3, pp.~1455--1468, 2020.

\bibitem{seliger2001flexible}
G.~Seliger, T.~Keil, U.~Rebafka, and A.~Stenzel, ``Flexible disassembly tools,'' in {\em Proceedings of the 2001 IEEE International Symposium on Electronics and the Environment. 2001 IEEE ISEE (Cat. No. 01CH37190)}, pp.~30--35, IEEE, 2001.

\bibitem{zuo2002novel}
B.-R. Zuo, A.~Stenzel, and G.~Seliger, ``A novel disassembly tool with screwnail endeffectors,'' {\em Journal of Intelligent Manufacturing}, vol.~13, pp.~157--163, 2002.

\bibitem{haldar2024baku}
S.~Haldar, Z.~Peng, and L.~Pinto, ``Baku: An efficient transformer for multi-task policy learning,'' {\em arXiv preprint arXiv:2406.07539}, 2024.

\bibitem{kumar2020conservative}
A.~Kumar, A.~Zhou, G.~Tucker, and S.~Levine, ``Conservative q-learning for offline reinforcement learning,'' {\em Advances in Neural Information Processing Systems}, vol.~33, pp.~1179--1191, 2020.

\bibitem{kostrikov2021offline}
I.~Kostrikov, A.~Nair, and S.~Levine, ``Offline reinforcement learning with implicit q-learning,'' {\em arXiv preprint arXiv:2110.06169}, 2021.

\bibitem{zare2024survey}
M.~Zare, P.~M. Kebria, A.~Khosravi, and S.~Nahavandi, ``A survey of imitation learning: Algorithms, recent developments, and challenges,'' {\em IEEE Transactions on Cybernetics}, 2024.

\bibitem{ross2011reduction}
S.~Ross, G.~Gordon, and D.~Bagnell, ``A reduction of imitation learning and structured prediction to no-regret online learning,'' in {\em Proceedings of the fourteenth international conference on artificial intelligence and statistics}, pp.~627--635, JMLR Workshop and Conference Proceedings, 2011.

\bibitem{florence2019self}
P.~Florence, L.~Manuelli, and R.~Tedrake, ``Self-supervised correspondence in visuomotor policy learning,'' {\em IEEE Robotics and Automation Letters}, vol.~5, no.~2, pp.~492--499, 2019.

\bibitem{pearce2023imitating}
T.~Pearce, T.~Rashid, A.~Kanervisto, D.~Bignell, M.~Sun, R.~Georgescu, S.~V. Macua, S.~Z. Tan, I.~Momennejad, K.~Hofmann, {\em et~al.}, ``Imitating human behaviour with diffusion models,'' {\em arXiv preprint arXiv:2301.10677}, 2023.

\bibitem{chi2023diffusion}
C.~Chi, Z.~Xu, S.~Feng, E.~Cousineau, Y.~Du, B.~Burchfiel, R.~Tedrake, and S.~Song, ``Diffusion policy: Visuomotor policy learning via action diffusion,'' {\em The International Journal of Robotics Research}, p.~02783649241273668, 2023.

\bibitem{lin2024learning}
T.~Lin, Y.~Zhang, Q.~Li, H.~Qi, B.~Yi, S.~Levine, and J.~Malik, ``Learning visuotactile skills with two multifingered hands,'' {\em arXiv preprint arXiv:2404.16823}, 2024.

\bibitem{pattabiraman2024learning}
V.~Pattabiraman, Y.~Cao, S.~Haldar, L.~Pinto, and R.~Bhirangi, ``Learning precise, contact-rich manipulation through uncalibrated tactile skins,'' {\em arXiv preprint arXiv:2410.17246}, 2024.

\bibitem{zhao2023learning}
T.~Z. Zhao, V.~Kumar, S.~Levine, and C.~Finn, ``Learning fine-grained bimanual manipulation with low-cost hardware,'' {\em arXiv preprint arXiv:2304.13705}, 2023.

\bibitem{he2016deep}
K.~He, X.~Zhang, S.~Ren, and J.~Sun, ``Deep residual learning for image recognition,'' in {\em Proceedings of the IEEE conference on computer vision and pattern recognition}, pp.~770--778, 2016.

\bibitem{ioffe2015batch}
S.~Ioffe, ``Batch normalization: Accelerating deep network training by reducing internal covariate shift,'' {\em arXiv preprint arXiv:1502.03167}, 2015.

\bibitem{wu2018group}
Y.~Wu and K.~He, ``Group normalization,'' in {\em Proceedings of the European conference on computer vision (ECCV)}, pp.~3--19, 2018.

\bibitem{ho2020denoising}
J.~Ho, A.~Jain, and P.~Abbeel, ``Denoising diffusion probabilistic models,'' {\em Advances in neural information processing systems}, vol.~33, pp.~6840--6851, 2020.

\bibitem{mandlekar2021matters}
A.~Mandlekar, D.~Xu, J.~Wong, S.~Nasiriany, C.~Wang, R.~Kulkarni, L.~Fei-Fei, S.~Savarese, Y.~Zhu, and R.~Mart{\'\i}n-Mart{\'\i}n, ``What matters in learning from offline human demonstrations for robot manipulation,'' {\em arXiv preprint arXiv:2108.03298}, 2021.

\end{thebibliography}
